\documentclass[10pt]{article}

\usepackage{iftex}
\ifPDFTeX
  \usepackage[utf8]{inputenc}
  \usepackage[T1]{fontenc}
  \usepackage{lmodern}
\else
  \usepackage{fontspec}
  \newfontfamily\arabicfont[
    Path = fonts/,
    Script = Arabic,
    UprightFont = Amiri-Regular.ttf,
    BoldFont = Amiri-Bold.ttf,
    ItalicFont = Amiri-Italic.ttf,
    BoldItalicFont = Amiri-BoldItalic.ttf
  ]{Amiri-Regular.ttf}
\fi
\usepackage{microtype}
\newcommand{\arabictext}[1]{\ifPDFTeX\textit{Arabic phrase}\else{\arabicfont #1}\fi}

\usepackage[
  top=2.5cm, bottom=2.8cm,
  left=2.5cm, right=2.5cm
]{geometry}

\usepackage{amsmath}
\usepackage{amssymb}

\usepackage{graphicx}
\usepackage{float}
\usepackage[
  labelfont=bf, font=small,
  justification=justified, singlelinecheck=false
]{caption}

\usepackage{booktabs}
\usepackage{array}
\usepackage[table]{xcolor}
\usepackage{tabularx}
\newcolumntype{C}{>{\centering\arraybackslash}X}
\newcolumntype{L}{>{\raggedright\arraybackslash}X}

\definecolor{brandcolor}{HTML}{000000}
\definecolor{lightlinecolor}{HTML}{CCCCCC}
\definecolor{accentgray}{HTML}{000000}
\definecolor{footergray}{HTML}{000000}
\definecolor{rowgray}{HTML}{F5F5F5}

\usepackage[
  colorlinks=true,
  linkcolor=black,
  citecolor=black,
  urlcolor=black
]{hyperref}

\usepackage{fancyhdr}
\usepackage{titlesec}

\titleformat{\section}
  {\normalfont\large\bfseries}{\thesection.}{0.5em}{}
\titlespacing{\section}{0pt}{18pt plus 3pt minus 2pt}{6pt}

\titleformat{\subsection}
  {\normalfont\normalsize\bfseries}{\thesubsection.}{0.5em}{}
\titlespacing{\subsection}{0pt}{12pt plus 2pt minus 1pt}{4pt}

\titleformat{\subsubsection}[runin]
  {\normalfont\normalsize\bfseries}{}{0em}{}[.\quad]
\titlespacing{\subsubsection}{0pt}{8pt plus 1pt}{0pt}

\usepackage{enumitem}
\setlist[itemize]{leftmargin=1.5em, topsep=3pt, itemsep=2pt, parsep=0pt}
\setlist[enumerate]{leftmargin=1.5em, topsep=3pt, itemsep=2pt, parsep=0pt}

\fancypagestyle{firstpage}{
  \fancyhf{}
  
  \fancyfoot[L]{\footnotesize\textcolor{footergray}{%
    \textcopyright~\the\year~Perle.ai. All rights reserved.}}
  \fancyfoot[R]{\small\textcolor{footergray}{1}}
}

\renewenvironment{abstract}{%
  \small\bfseries\noindent\ignorespaces
}{\par\medskip}

\newcommand{\brandrule}{%
  \noindent\textcolor{brandcolor}{\rule{\linewidth}{1.5pt}}\par%
}
\newcommand{\lightrule}{%
  \noindent\textcolor{lightlinecolor}{\rule{\linewidth}{0.4pt}}\par%
}

\begin{document}
\thispagestyle{firstpage}

\begin{flushleft}
  \includegraphics[height=20pt]{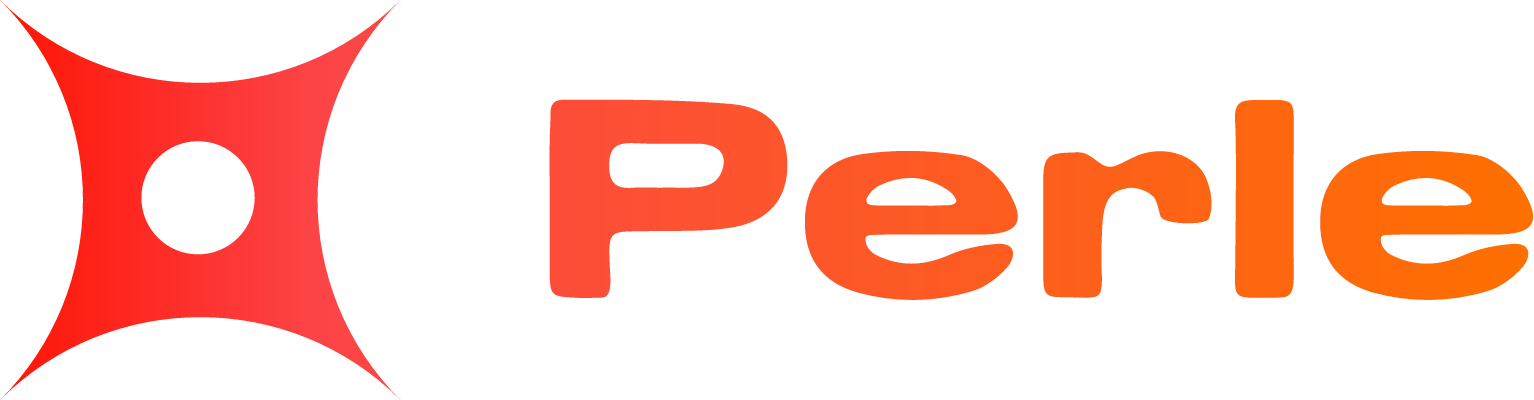}%
  \hfill
  \textcolor{accentgray}{\small\today}
\end{flushleft}

\vspace{4pt}
\brandrule
\vspace{8pt}

\noindent{\LARGE\bfseries Benchmarking Frontier LLMs on Arabic Cultural and Sociolinguistic Knowledge: A Cross-Evaluation Framework with Human SME Ground Truth \par}

\vspace{10pt}

\noindent
\textbf{Sajjad Abdoli}\textsuperscript{1,*,\dag}\footnote{Corresponding author: \href{mailto:sajjad@perle.ai}{\texttt{sajjad@perle.ai}}},\quad
\textbf{Ghassan Al-Sumaidaee}\textsuperscript{1,*,\dag}\footnote{Corresponding author: \href{mailto:ghassan.al-sumaidaee@perle.ai}{\texttt{ghassan.al-sumaidaee@perle.ai}}; ORCID: \href{https://orcid.org/0000-0002-5536-0252}{0000-0002-5536-0252}},\quad
\textbf{Ahmad ElShiekh}\textsuperscript{1,*}\footnote{ORCID: \href{https://orcid.org/0009-0001-6837-6202}{0009-0001-6837-6202}},\quad
\textbf{Clayton W. Taylor}\textsuperscript{1,*}\footnote{ORCID: \href{https://orcid.org/0009-0006-6478-8994}{0009-0006-6478-8994}},\quad
\textbf{Ahmed Rashad}\textsuperscript{1}
\par\vspace{3pt}
{\small\textsuperscript{1}Perle AI\quad
\textsuperscript{*}Equal contribution; names sorted alphabetically.\quad
\textsuperscript{\dag}Corresponding authors\\[2pt]
\texttt{sajjad@perle.ai}\quad
\texttt{ghassan.al-sumaidaee@perle.ai}\quad
\texttt{clayton@perle.ai}\\[2pt]
\texttt{mad.elshiekh@perle.ai}\quad
\texttt{ahmed@perle.ai}}

\vspace{10pt}

{
\begin{abstract}The cost of human expert evaluation is a principal bottleneck to deploying
language models in specialized, {often high-stakes domains. This bottleneck is particularly acute regarding Arabic
sociolinguistic knowledge}: credible grading requires not only
linguistic fluency but deep cultural familiarity that cannot be approximated by
surface-level, {black-and-white metrics and training data}.
 { We address this with a cross-evaluation
framework instantiated on two underrepresented Arabic dialect communities:
Egyptian and Iraqi Arabic. We contribute a dataset of 103 validated
prompt--rubric pairs (70 Egyptian and 33 Iraqi; 53 of which are Cultural, the other 50 Linguistic),
authored and graded by native-speaker SMEs using penalty-weighted rubrics
that distinguish positive content requirements from answer-specific negative error criteria. Three frontier LLMs serve as \emph{target models} (whose responses are graded by human SMEs across 302 unique prompt--response pairs), while five frontier LLMs serve as \emph{automated judges} grading target model responses against the rubric, enforcing a provider-level self-evaluation guard. A dual-metric
scheme combining Mean Absolute Deviation (MAD) with a Signed Mean Error
separates directional grading bias from symmetric noise. Across 1,307 judge
evaluations: GPT-5.4 is the most reliable judge ($\mathrm{MAD}_j = 10.21$~pp,
Signed Error $= {-1.12\%}$); four of five judges show systematic leniency
($+2.01\%$ to $+6.56\%$); Cultural tasks are harder to grade than Linguistics
tasks for all judges (MAD gap $1.83$--$4.78$~pp); { and models substantially outperform on Egyptian prompts {
 when compared to outputs for Iraqi prompts . However, given the difference in leniency between the Iraqi and Egyptian SMEs scoring model outputs, we cannot solely attribute the Egyptian-Iraqi performance gap to model knowledge alone. We have therefore chosen to emphasize findings that do not assume identical leniency across human graders. Across all samples, regardless of subdomain or judge leniency, implicit
cultural reasoning, which requires models simulate native-speaker judgment rather
than rely on lexical verification,} emerges as the primary failure mode for
automated grading across all judge models.}}
\end{abstract}}

\vspace{6pt}
\lightrule
\vspace{14pt}

\section{Introduction}

{
Objective, verifiable domains like math and science, in which experts generally agree on what distinguishes ``true'' from ``false,'' are well-suited for reinforcement learning (``RL'') with binary reward signals. Language, however, and the experiences we use it to describe, are intrinsically subjective. Subjective domains (e.g. belief systems, cultures, linguistic dialects and subdialects) therefore elude binary benchmarking techniques employed for more ``objective,'' verifiable domains. This necessitates the use of a limited pool of human subject matter experts (``SMEs'') for model evaluation and feedback, which in turn embed certain bottlenecks in RL environments themselves. The adaptation of techniques that've helped scale feedback and reward signaling in more ``objective,'' verifiable domains, such as LLMs-as-judges, is therefore essential to the advancement of AI in those that are inherently subjective.
The promise of LLM-as-a-Judge frameworks~\cite{llm_as_judge} is that a capable
frontier model can substitute for a human evaluator, drastically reducing annotation
costs. Embedded in this promise, however, is the risk of systematic bias: a model that is too lenient underestimates error
rates and misleads downstream quality decisions; one that is too conservative
suppresses signal on genuinely correct outputs. On rubrics dominated by penalty
criteria, such as nuanced cultural queries where errors are more common than
successes, leniency and high deviation from the human baseline are the same
underlying phenomenon, not independent failures.}

In this paper we ask: \textit{which frontier LLMs can reliably substitute for human
SMEs when grading Arabic sociolinguistic tasks, and where do they
systematically fail?} We answer this through a cross-evaluation framework. Every
frontier model is evaluated as both a \emph{target} (its answers are graded by human
SMEs, establishing a gold standard) and as a \emph{judge} (it grades the answers of
all other models, never its own). This yields a complete grading matrix from which we
derive judge-level reliability measures and identify the specific criterion types where
automated and human grading diverge.

Our contributions are:

\begin{itemize}
  \item A cross-evaluation benchmark framework for Arabic cultural and sociolinguistic
        grading, with a provider-level self-evaluation guard that prevents any model
        from judging outputs from the same provider, regardless of model version.
  \item A dataset of 103 validated prompt--rubric pairs across Cultural and Linguistics
        domains (Egyptian Arabic: {70 prompts}; Iraqi Arabic: {33 prompts}),
        authored and graded by domain-expert SMEs using a structured annotation platform with
        explicit positive and negative criteria.
  \item Evidence that GPT-5.4 is the most reliable automated judge
        ($\mathrm{MAD}_j = 10.21$~pp, near-zero directional bias), while all other
        judges show systematic leniency bias, with Cultural tasks harder to grade
        than Linguistics tasks for all five judges.
  \item A Signed Mean Error metric that separates directional grading bias (leniency
        vs.\ conservatism) from symmetric noise, enabling more actionable
        characterization of each judge model's failure mode.
\end{itemize}

The paper is organized as follows. Section~\ref{sec:background} situates this work
in the LLM-as-a-Judge literature. Section~\ref{sec:dataset} describes the dataset
and annotation platform. Section~\ref{sec:methodology} formalizes the cross-evaluation
framework and metrics. Section~\ref{sec:results} reports results.
Section~\ref{sec:discussion} interprets the findings.
Section~\ref{sec:conclusion} concludes.

{
\section{Background and Motivation}
\label{sec:background}

\subsection{The Challenge of Culturally Specific LLMs}

The deployment of large language models in culturally sensitive, non-Western contexts
exposes a fundamental tension: language fluency and cultural alignment are distinct
capabilities, and contemporary LLMs systematically conflate them.
Hershcovich et al.~\cite{hershcovich2022challenges} first formalized this distinction,
showing that speakers vary not only by language but by culture-specific aboutness,
common ground, values, and linguistic register --- and that standard NLP evaluation
instruments designed for English largely fail to capture these dimensions.
Adilazuarda et al.~\cite{adilazuarda2024survey} surveyed more than 90 subsequent papers
and found that \textit{none} explicitly define ``culture''; instead, they probe models
via proxy datasets representing selected cultural facets, leaving semantic domains and
real-world application effects largely under-studied.

A growing body of work has demonstrated that the root cause of cultural misalignment
lies upstream in the training pipeline.
Sahu et al.~\cite{sahu2026culturefunnel} diagnose this as a \emph{cultural data funnel}:
explicit cultural signals decline sharply across fine-tuning, alignment, and reasoning
stages, while geographically concentrated, task-specialized data dominates.
Their analysis, grounded in a 5.6M-sample culturally tagged corpus spanning 194
languages, confirms that multilinguality enhances geographic diversity of cultural
knowledge but does not ensure balanced representation.
Agarwal et al.~\cite{agarwal2026fluentbutforeign} evaluate Indic and global LLMs
against nationally representative surveys and community-sourced QA in India and find
that regional LLMs do not align better with local norms than global ones ---
attributing the failure to scarce culturally grounded pretraining data and
demonstrating that prompting and regional fine-tuning cannot recover alignment and can
even degrade existing cultural knowledge.

The multilingual--multicultural gap is equally pronounced across other language families.
Rystr{\o}m et al.~\cite{ryostrom2025multilingual} compare LLM response distributions
against World Values Survey population data across Danish, Dutch, English, and
Portuguese and find no consistent relationship between multilingual capability and
cultural alignment across model families, establishing self-consistency rather than
language capability as a stronger predictor of multicultural alignment.
The broader landscape of cultural misrepresentation in LLMs is surveyed in
Shi et al.~\cite{shi2024culturebank}, who construct CultureBank from 12K TikTok and
11K Reddit self-narratives to provide a community-driven cultural knowledge base, and
show that fine-tuning on this resource improves performance on downstream cultural
tasks in zero-shot settings.

For the Arabic-speaking world specifically, the challenge is compounded by diglossia
between Modern Standard Arabic and spoken varieties, dialectal variation across
22 Arab countries spanning over 450 million speakers, and the substantial presence
of machine-translated rather than natively authored training data.
Alwajih et al.~\cite{alwajih2025palmx} argue that many Arabic LLMs, despite achieving
linguistic fluency, largely overlook country-specific cultural competence because they
are trained on machine-translated datasets and evaluated on general NLP tasks.
Their PalmX 2025 shared task provides the first standardized benchmark for cultural
competence across Arabic and Islamic domains.
Qian et al.~\cite{qian2024cameleval} introduce CamelEval, an LLM-as-a-Judge benchmark
for culturally aligned Arabic models, and demonstrate systematic gaps between
Arabic-specific instruction-following quality and genuine cultural understanding.
Zhang et al.~\cite{zhang2026mindthegap} propose CultureManager, a modular pipeline
for task-specific cultural alignment that stores multi-culture knowledge in separate
adapters to avoid cross-culture interference, and show that both the relevance of
cultural norms to the task at hand and interference between cultures are distinct
failure modes requiring architectural solutions rather than inference-time workarounds.
Oh et al.~\cite{oh2025cultureiseverywhere} argue further that cultural assumptions
permeate even ostensibly neutral evaluations, and call for intentional cultural
evaluation that systematically examines these assumptions in \textit{all} benchmark
design, not only in explicitly cultural tasks.

\subsection{Evaluating LLMs: Methods, Benchmarks, and Cultural Dimensions}

The evaluation of LLM outputs has evolved across three broad paradigms:
\emph{direct preference assessment}, where a judge rates a single output on a numeric
scale; \emph{pairwise ranking}, where a judge selects the better of two responses; and
\emph{rubric-based evaluation}, where quality is decomposed into independently
assessable binary criteria.
Zheng et al.~\cite{llm_as_judge,mtbench} establish LLM-as-a-Judge as a scalable
alternative to human annotation, demonstrating strong correlation with human preferences
for instruction-following and conversational quality.
However, systematic biases have been documented in all three paradigms: positional bias
(favouring answers in certain positions), verbosity bias (favouring longer responses),
and self-enhancement bias (favouring outputs from the same model family)~\cite{llm_as_judge}.
Fu and Liu~\cite{fu2025multilingual_judge} show that these biases are further amplified
in multilingual settings: LLM-as-a-Judge reliability degrades substantially for
non-English languages, motivating the native-speaker SME ground truth used in the
present work.

Cultural dimensions introduce further complications for each evaluation paradigm.
Parrish et al.~\cite{parrish2022bbq} establish BBQ, a 58,492-example hand-built bias
benchmark covering nine U.S.-anchored social dimensions, demonstrating that models
rely on stereotypes under under-informative context and retain accuracy advantages
when correct answers align with social biases.
The U.S.-centric framing of BBQ exemplifies the critique by Oh et al.~\cite{oh2025cultureiseverywhere}
that evaluation design itself encodes cultural assumptions.
Huang and Yang~\cite{huang2023culturenli} operationalize cultural variation as label
disagreement in natural language inference, introducing CALI, a 2.7K dataset annotated
by U.S. and Indian cultural groups, and show that models default to culture-generic
labels rather than exhibiting genuine cultural bias --- a nuance that aggregate metrics
cannot capture.

Cultural benchmark design has matured substantially in recent years.
Koto et al.~\cite{koto2024blend} introduce BLEnD, a hand-crafted benchmark of 52.6K
everyday cultural QA pairs from 16 countries in 13 languages, and document a maximum
57.34 percentage point performance gap between cultures that are more versus less
represented online --- the strongest existing quantitative link between data presence
and cultural performance.
Rao et al.~\cite{rao2025normad} introduce NormAd, a hierarchical framework spanning
2.6K social-etiquette scenarios from 75 countries, and find that even the best LLMs
achieve less than 82\% accuracy when the relevant social norms are provided as context,
compared with over 95\% for humans --- a gap that widens markedly for Global South
cultures.
Zhang et al.~\cite{zhang2025culturescope} introduce CultureScope, a three-layer
dimensional schema (Institutional Norms; Behavioral Patterns; Core Values and Social
Structures) across 140 dimensions, enabling automated construction of culture-specific
evaluation sets for any language and confirming that multilingual data does not
necessarily enhance cultural understanding.
Hayes Zhang et al.~\cite{hayeszhang2025community} demonstrate that human preferences
are substantially more diverse than model outputs, and introduce the Community
Alignment Dataset comprising 233,319 comparisons across five countries, establishing
that candidate-response homogeneity prevents reward models from learning heterogeneous
and underserved preferences.

In the context of rubric-based evaluation specifically, Kim et al.~\cite{kim2024prometheus}
pioneer Prometheus, a fully open-source LLM evaluator trained on 1K fine-grained score
rubrics and 100K natural language feedback instances, demonstrating that open-source
models can match GPT-4 evaluation capabilities when appropriate rubric structure and
reference materials are provided.
This work establishes the rubric-based paradigm as practically viable and reproducible
at scale, and is extended in Prometheus~2~\cite{kim2024prometheus2}, which supports
both direct assessment and pairwise ranking under user-defined evaluation criteria.

\subsection{Rubric Design for Grounded Evaluation}

The reliability of rubric-based evaluation depends critically on the quality of the
criteria themselves. Verifiable reward methods have scaled reinforcement learning in
objective domains by converting task success into checkable binary signals~\cite{su2025rewardbridge}; rubric-based evaluation attempts to recover an analogous signal
structure for domains where correctness is culturally or linguistically situated.
Zhang et al.~\cite{zhang2026chasingtail} provide the theoretical foundation: reward
over-optimization in LLM post-training ``primarily arises from inaccuracy in
high-reward regions,'' and rubric-based rewards mitigate this by distributing the
evaluation signal across independently verifiable dimensions.
Their core empirical finding is that effective rubric construction requires two
properties: (1) criteria must distinguish \emph{excellent from merely great}
responses, not just correct from incorrect; and (2) rubrics must discriminate among
\emph{diverse off-policy responses}, not only the on-policy distribution.
Li et al.~\cite{li2026rubrichub} extend this by introducing RubricHub, a large-scale
collection of coarse-to-fine generated rubrics showing that criterion granularity and
discriminability are the primary determinants of reward model accuracy in post-training.

These theoretical principles have clear practical counterparts.
Kim et al.~\cite{kim2024prometheus} and Kim et al.~\cite{kim2024prometheus2} show that
criterion clarity --- the degree to which a rubric item can be assessed without
ambiguity --- is the primary determinant of inter-judge reliability.
Operationalizing rubric quality for domain-expert SME-authored criteria in our setting,
we instruct annotators to follow six structural principles derived from the rubric
design literature~\cite{zhang2026chasingtail,kim2024prometheus,li2026rubrichub}:

\begin{enumerate}[leftmargin=2em, topsep=4pt, itemsep=4pt, parsep=0pt]

  \item \textbf{Mutually Exclusive and Collectively Exhaustive (MECE).}
        Criteria must not overlap --- the same error in a model's response must not
        be penalised by two separate criteria --- and together the criteria must cover
        all aspects of a perfect response with no gaps.

  \item \textbf{Complete.}
        The rubric as a whole must be sufficient to evaluate any response to the
        prompt; a response satisfying all positive criteria and triggering no negative
        criteria must constitute a genuinely high-quality answer.

  \item \textbf{Non-overlapping.}
        Each factual claim or cultural property appears in exactly one criterion;
        redundancy inflates the penalty for a single error and distorts the score
        distribution.

  \item \textbf{Atomic / Non-stacked.}
        Each criterion evaluates exactly one distinct aspect. Compound criteria
        combining multiple conditions with ``and'' are split into separate items,
        guaranteeing that binary PASS/FAIL evaluation is unambiguous.

  \item \textbf{Self-contained.}
        Each criterion contains all information needed to evaluate a response without
        requiring external reference. A criterion that names a fact must embed the
        fact: \emph{not} ``Mentions the capital of Canada'' but ``Mentions that the
        capital of Canada is Ottawa.''

  \item \textbf{Verifiable without external search.}
        The judge must assess each criterion solely from the rubric text and the
        model's response. Criteria requiring enumeration of acceptable answers must
        list them: \emph{not} ``Names a Nobel Prize winner in Physics in 2023'' but
        ``Names at least one of the following Nobel Prize winners in Physics in 2023:
        Pierre Agostini, Ferenc Krausz, or Anne L'Huillier.''

\end{enumerate}

These constraints collectively address the principal rubric failure modes documented
in the literature~\cite{zhang2026chasingtail,kim2024prometheus,li2026rubrichub}:
criterion overlap (double-penalization), ambiguity (divergent judge interpretations),
and implicit facts (criteria that no automated judge can verify from the rubric text
alone).
The self-contained and verifiable-without-search requirements carry additional weight
in the cultural domain: a culturally implicit criterion --- one encoding native-speaker
knowledge --- cannot be fairly assessed by any judge unless the knowledge is made
explicit within the criterion itself.
This design choice is directly connected to the Explicit versus Implicit dimension
used to classify rubric criteria throughout our analysis
(Section~\ref{sec:dataset:rubric}), and explains the systematic judge--SME divergence
we observe on Implicit Cultural criteria in Section~\ref{sec:results}.

\subsection{Penalty-Weighted Rubrics and Their Properties}

A distinctive feature of our evaluation setting is that the rubric is dominated by
\emph{negative criteria} --- penalty-weighted items that are triggered when the target
model makes a cultural or factual error. This produces a ground-truth baseline that
is typically negative (i.e., the sum of penalties exceeds positive credits), reflecting
the frequency of errors in LLM responses on culturally demanding tasks. A lenient
judge that fails to trigger these penalties will score closer to zero than the SME,
mechanically producing a higher Mean Absolute Deviation. This property means that on
our rubrics, leniency bias and grading deviation are not independent findings --- they
are the same underlying failure measured in two ways.
}

\section{Dataset}
\label{sec:dataset}

\subsection{SME-Authored Prompts}
\label{sec:dataset:prompts}

Prompts and ground-truth grades are produced by a panel of domain-expert SMEs
drawn from our in-house expert network. Each SME is a native speaker of the dialect
community they author for. SMEs are recruited through a structured process:
credential verification, a screening stage, and a comprehensive onboarding protocol
covering the annotation platform and the criterion taxonomy.

In the current dataset, each (target model, sample) pair is graded by a single SME
from the relevant dialect community. This single-annotator design means the ground
truth carries unmeasured annotator noise; inter-annotator agreement is the empirical
ceiling for any automated judge and will be quantified in future work.

{
\subsubsection{Two-Phase Authoring and Scoring Workflow}
\label{sec:dataset:workflow}

Prompts and rubrics are produced through a two-phase workflow that separates
model-agnostic design from model-specific scoring, and this separation holds
identically for both dialect communities and both domains.

\textbf{Phase A (pre-output)} is performed before any target model is run. The author
fixes the scope (Cultural or Linguistics, never mixed within a single rubric), drafts
the prompt in the target dialect, validates that the prompt is genuinely difficult by
submitting it once to a model in a clean session and discarding that throwaway output,
and then builds the positive criteria directly from the SME gold answer. Because
positive criteria are derived from the ground truth rather than from any observed
response, they are model-agnostic and are reused unchanged whenever a new target is
evaluated on the same prompt. This difficulty-validation step is also the reason the
SME baseline is low and penalty-heavy (Section~2.4): prompts a model answers perfectly
in validation are revised or discarded before a rubric is built.

\textbf{Phase B (post-output)} is repeated for every target model. The response is
first scored against the fixed positive criteria as PASS or FAIL, after which the
author adds answer-specific negative criteria for the errors actually present in that
response, each tagged with one failure type (Section~\ref{sec:dataset:failure-taxonomy}). Positive criteria thus
remain constant across targets for a given prompt, while the set and weights of
negative criteria vary by response. The two roles, rubric author (Phase A) and output
scorer (Phase B), may be the same SME or staffed separately; in the current dataset a
single SME per (sample, target) pair performs both, which is the source of the
single-annotator noise discussed in Section~6.5.
}

\subsection{Rubric Design}
\label{sec:dataset:rubric}

Each rubric consists of two criterion types:

\textbf{Positive criteria} are shared across all responses to a given prompt. They
represent factual or cultural elements a correct answer must contain, each carrying
a positive weight. A judge assigns PASS if the criterion is satisfied, FAIL otherwise.

\textbf{Negative criteria} are answer-specific: they are populated by the SME only
for errors actually observed in a given model's response, each carrying a negative
weight. A judge assigns ``Error Present'' if the described error is present, ``0''
otherwise. Because negative criteria reflect observed errors rather than a static
checklist, their count and weights vary across responses to the same prompt.

{
\subsubsection{Negative-Criterion Failure Taxonomy}
\label{sec:dataset:failure-taxonomy}

Every negative criterion authored by an SME is tagged with one of six failure types,
enabling error-mode analysis beyond a raw penalty count. The taxonomy is:
\textsc{Hallucination}: the model fabricates a term, fact, proverb, or explanation
that does not exist (e.g.\ inventing the saying ``\arabictext{جسر بلا باب مثل شكر بلا چاي}'' and
attributing it to Iraqi popular culture); \textsc{Omission}: the model fails to
provide content the criterion explicitly requires (e.g.\ never identifying ``\arabictext{حمرة}''
and ``\arabictext{خضرة}'' as the colour-based colloquial names for the 25,000 and 10,000 dinar
notes); \textsc{Cultural Misattribution}: a pan-Arabic, Gulf, Levantine, or otherwise
non-Iraqi term or concept is presented as specifically Iraqi, or an existing Iraqi
term is assigned to the wrong referent (e.g.\ presenting the Gulf greeting
``\arabictext{ ومرحبا والله هلا}'' as the defining Iraqi guest greeting, or defining ``\arabictext{التشريبة}''
as eating dates with tea rather than the bread-and-broth dish); \textsc{Ambiguous
Framing}: non-native metaphors, dialect-register inconsistency, or misleading
phrasing (e.g.\ using the Levantine ``\arabictext{شو}'' in place of Iraqi ``\arabictext{شنو}'', or the
Jordanian ``\arabictext{سخن}'' for ``hot'' where Iraqi uses ``\arabictext{حار}''); \textsc{Irrelevant
Addition}: unsolicited content not requested by the user, which inflates surface
coverage while creating openings for further error; and \textsc{Rendering Failure}:
the model generates invalid characters in place of Arabic script. This taxonomy is
the unit at which we diagnose why automated judges and SMEs diverge
(Section~6.3): a judge may correctly fire an \textsc{Omission}
penalty it can verify lexically while missing a \textsc{Cultural Misattribution}
that requires native dialect knowledge.
}

{
\subsubsection{Worked Example (Iraqi Cultural)}
\label{sec:dataset:worked-example}

Consider the Iraqi prompt asking for the colloquial names Iraqis use for the 25,000
and 10,000 dinar banknotes. The gold answer is ``\arabictext{حمرة}'' (from the red
note) and ``\arabictext{خضرة}'' (from the green note), with the naming convention
deriving from the physical colour of each note. Positive criteria are shared across
all target responses to this prompt; negative criteria are populated per response
from the errors an SME actually observes. Table~\ref{tab:iraqi_banknote_rubric}
shows the full rubric for one target response (Muse Spark), which scored $-26$
against a maximum of 37 ($-70.3\%$). Criterion weights reflect importance bands
applied consistently across the Iraqi subset: core term identification 8--10,
meaning and context 5--9, cultural depth 3--7, and dialect and register 3--5.

\setcounter{table}{0}
\begin{table}[H]
\centering
\caption{Complete rubric and SME verdicts for one Muse Spark response to CULT-MENA-IRQ-N. Positive criteria (max 37) are shared across targets; answer-specific negative penalties yield a raw score of $-26$ ($-70.3\%$), illustrating the taxonomy in Section~\ref{sec:dataset:failure-taxonomy}.}
\label{tab:iraqi_banknote_rubric}
\scriptsize
\renewcommand{\arraystretch}{1.28}
\setlength{\tabcolsep}{3pt}
\begin{tabularx}{\textwidth}{>{\raggedright\arraybackslash}p{0.10\textwidth} L >{\raggedright\arraybackslash}p{0.10\textwidth} >{\centering\arraybackslash}p{0.08\textwidth} >{\raggedright\arraybackslash}p{0.18\textwidth} L}
  \toprule
  \textbf{ID} & \textbf{Criterion (abridged)} & \textbf{Type} & \textbf{Weight} & \textbf{Dimension / Explicitness} & \textbf{Verdict / Failure} \\
  \midrule
  \rowcolor{rowgray}
  IRQ-6-1 & Identifies ``\arabictext{حمرة}'' for the 25,000 note (red colour) & Positive & $+10$ & Accuracy / Explicit & FAIL / \textsc{Omission} \\
  IRQ-6-2 & Identifies ``\arabictext{خضرة}'' for the 10,000 note (green colour) & Positive & $+10$ & Accuracy / Explicit & FAIL / \textsc{Omission} \\
  \rowcolor{rowgray}
  IRQ-6-3 & Explains names derive from physical banknote colour & Positive & $+8$ & Accuracy / Explicit & FAIL / \textsc{Omission} \\
  IRQ-6-4 & Stays focused on the two denominations asked about & Positive & $+5$ & Completeness / Explicit & FAIL / \textsc{Irrelevant Addition} \\
  \rowcolor{rowgray}
  IRQ-6-5 & Uses authentic Iraqi dialect and natural phrasing & Positive & $+4$ & Communication Quality / Implicit & FAIL / \textsc{Ambiguous Framing} \\
  IRQ-6-6 & Presents ``\arabictext{ربع}'' (= 250 dinars) as the name for 25,000 & Negative & $-6$ & Accuracy / Explicit & Error Present / \textsc{Cultural Misattribution} \\
  \rowcolor{rowgray}
  IRQ-6-7 & Presents ``\arabictext{ورقات عشر}'' (= 1,000 USD) as a name for 10,000 dinars & Negative & $-5$ & Accuracy / Explicit & Error Present / \textsc{Cultural Misattribution} \\
  IRQ-6-8 & Presents ``\arabictext{ورقة}'' (= 100 USD) as referring to 100,000 dinars & Negative & $-5$ & Accuracy / Explicit & Error Present / \textsc{Cultural Misattribution} \\
  \rowcolor{rowgray}
  IRQ-6-9 & Presents ``\arabictext{نص}'' as the colloquial name for 50,000 dinars & Negative & $-4$ & Accuracy / Explicit & Error Present / \textsc{Hallucination} \\
  IRQ-6-10 & Attributes the naming convention to post-2003 inflation & Negative & $-4$ & Accuracy / Explicit & Error Present / \textsc{Hallucination} \\
  \rowcolor{rowgray}
  IRQ-6-11 & Ends by soliciting more information not asked for & Negative & $-2$ & Communication Quality / Explicit & Error Present / \textsc{Ambiguous Framing} \\
  \bottomrule
\end{tabularx}
\end{table}
}

Both criterion types carry an \emph{Objectivity} dimension (Objective vs.\
Subjective) and an \emph{Explicitness} dimension (Explicit vs.\ Implicit), enabling
fine-grained analysis of where automated and human grading diverge. The LLM judge
receives only criterion text and structure --- never the SME's Pass/Fail labels.

\subsection{Dataset Scale and Dialect Breakdown}
\label{sec:dataset:scale}

Table~\ref{tab:split} shows the dataset distribution. The full dataset comprises
103 validated prompt--rubric pairs across two domains and two dialect communities.

\begin{table}[H]
\centering
\caption{Dataset distribution by dialect community and domain.}
\label{tab:split}
\small
\renewcommand{\arraystretch}{1.3}
\begin{tabularx}{\textwidth}{l C C C C}
  \toprule
  \textbf{Dialect} & \textbf{Code} & \textbf{Cultural} & \textbf{Linguistics} & \textbf{Total} \\
  \midrule
  \rowcolor{rowgray}
  Iraqi Arabic    & IRQ & 15 & 18 & 33 \\
  Egyptian Arabic & EGY & 38 & 32 & 70 \\
  \midrule
  \textbf{Total}  &     & \textbf{53} & \textbf{50} & \textbf{103} \\
  \bottomrule
\end{tabularx}
\end{table}

\section{Methodology}
\label{sec:methodology}

\subsection{The Cross-Evaluation Framework}
\label{sec:methodology:framework}

The benchmark runs every judge model against every target model's responses,
enforcing a \textbf{provider-level self-evaluation guard}: no judge from the same
provider as a target model evaluates that target's responses, regardless of model
version (e.g.\ \texttt{gemini-3.1-pro-\allowbreak preview} as a target correctly blocks
\texttt{gemini-3.1-pro-\allowbreak preview} as a judge, since a model cannot judge its own
provider's output). This eliminates self-enhancement bias more robustly than a
model-level guard.

\textbf{Target model.} The model whose answers are being assessed. For each target,
human SMEs independently grade its responses, producing the ground-truth scores.

\textbf{Judge model.} The LLM assigned to grade the target model's answers against
the rubric. Every judge evaluates all targets outside its provider; it never evaluates
outputs from the same provider. The judge receives only rubric criterion text and
structure --- not the SME's Pass/Fail labels.

Running all cross-provider combinations yields two orthogonal analytical views:
the \emph{judge-axis} (most reliable automated grader) and the
\emph{target-axis} (whose answers are hardest to grade consistently).

\subsection{The Evaluation Triad}
\label{sec:methodology:triad}

For each (target model $t$, sample $s$) pair, the judge model receives:

\begin{enumerate}
  \item \textbf{The Prompt.} A cultural or sociolinguistic question authored by a human SME.
  \item \textbf{The Answer.} The target model's response to that prompt.
  \item \textbf{The Rubric.} The criterion text, score type (Positive/Negative),
        and criterion weights. The judge does \emph{not} receive the SME's Pass/Fail labels.
\end{enumerate}

{The judge is invoked with a fixed system prompt that assigns the role of an Arabic
sociolinguistics SME and specifies the exact JSON output schema; the user message
supplies the prompt, the target model's answer, and the rubric as a plain-text
table (criterion text, score type, weight only --- SME verdicts withheld).
The complete prompt text is reproduced in Appendix~\ref{app:judge_prompt}. The judge returns a structured object with, for each criterion: a PASS/FAIL
judgment (Positive criteria) or ``Error Present / 0'' flag (Negative criteria), and
a free-text justification.}

\subsection{Models Under Study}
\label{sec:methodology:models}

Table~\ref{tab:models} lists all models. Muse Spark is evaluated as a target only
(no public API; responses entered manually).

\begin{table}[H]
\centering
\caption{Models under study. Muse Spark is target-only.}
\label{tab:models}
\small
\renewcommand{\arraystretch}{1.3}
\begin{tabularx}{\textwidth}{l l l l}
  \toprule
  \textbf{Model} & \textbf{Provider} & \textbf{API Identifier} & \textbf{Role} \\
  \midrule
  \rowcolor{rowgray}
  GPT-5.4        & OpenAI    & \texttt{gpt-5.4-2026-03-05}        & Judge + Target \\
  gemini-3.1-pro-preview & Google & \texttt{gemini-3.1-pro-\allowbreak preview} & Judge + Target \\
  \rowcolor{rowgray}
  Grok 4.3       & xAI       & \texttt{grok-4.3}                  & Judge + Target \\
  Claude Opus 4.7 & Anthropic & \texttt{claude-opus-4-7}           & Judge only \\
  \rowcolor{rowgray}
  Qwen Plus      & Alibaba   & \texttt{qwen3.7-plus}              & Judge only \\
  \midrule
  Muse Spark     & Meta      & \textit{N/A (manual entry)}        & Target only \\
  \bottomrule
\end{tabularx}
\end{table}

\noindent All judge calls use a system prompt instructing the model to act as an
Arabic sociolinguistics SME and return only a structured JSON evaluation.
Temperature is fixed to 0 to eliminate stochastic variation.

{
Target responses are collected under controlled conditions to ensure independence and
reproducibility. Each prompt is submitted in a clean session with memory and
conversation history disabled, one prompt per session with no chaining, and the first
response is captured verbatim with no re-prompting or follow-up turns. Muse Spark,
which has no public API, is run under the same controls with responses entered
manually.
}

\subsection{Evaluation Metrics}
\label{sec:methodology:metrics}

{
Let a response $r = (s,m)$ denote a (sample,
target-model) pair, and let $R$ be the set of all human-graded responses
($|R| = 302$). For a judge $j$, let $R_j \subseteq R$ be the responses $j$ is
permitted to grade under the provider-level guard (so $|R_j|$ equals the Eval Rows
in Table~\ref{tab:judge_summary}). $j$ indexes a judge model; $P(r,j)$ is the percentage score judge $j$
assigns to response $r$; $P_{\mathrm{sme}}(r)$ is the human SME score (ground truth)
for $r$.
}

\subsubsection{Max Score and Raw Score}

\begin{equation}
  S_{\max} = \sum_{i\,\in\,\text{Positive}} w_i
  \label{eq:smax}
\end{equation}

\begin{equation}
  S_{\mathrm{raw}}(r,j)
  = \sum_{\substack{i\,\in\,\text{Positive}\\\cap\;\text{PASS}}} w_i
  + \sum_{\substack{k\,\in\,\text{Negative}\\\cap\;\text{Error}}} w_k
  \label{eq:sraw}
\end{equation}

\subsubsection{Percentage Score}

\begin{equation}
  P(r,j) = \frac{S_{\mathrm{raw}}(r,j)}{S_{\max}} \times 100
  \label{eq:pct}
\end{equation}

Because the rubric is dominated by Negative criteria ($w_k < 0$), $P_{\mathrm{sme}}(r)$
is often negative. A lenient judge scoring closer to zero than the SME mechanically
produces a higher MAD --- leniency bias and grading deviation are the same phenomenon.

\subsubsection{Judge-Axis MAD}

{
\begin{equation}
  \mathrm{MAD}_j
  = \frac{1}{|R_j|} \sum_{r\,\in\,R_j} \bigl|P(r,j) - P_{\mathrm{sme}}(r)\bigr|
  \label{eq:madj}
\end{equation}
}

\subsubsection{Target-Axis MAD}

{

For a fixed target model $m$, averaging over the cross-provider judges that graded
its responses, let $R_{m,j}$ be those evaluated response--judge pairs and let $N$
be their count:
\begin{equation}
  \mathrm{MAD}_t(m)
  = \frac{1}{N} \sum_{r\,\in\,R_{m,j}} \bigl|P(r,j) - P_{\mathrm{sme}}(r)\bigr|
  \label{eq:madt}
\end{equation}
}

\subsubsection{Signed Mean Error --- Bias vs.\ Noise}

{
\begin{equation}
  \mathrm{SME}_{\mathrm{err}}(j)
  = \frac{1}{|R_j|} \sum_{r\,\in\,R_j} \bigl(P(r,j) - P_{\mathrm{sme}}(r)\bigr)
  \label{eq:sme_err}
\end{equation}
}

A positive $\mathrm{SME}_{\mathrm{err}}$ indicates leniency; negative indicates
conservatism. A judge can have low MAD but non-zero $\mathrm{SME}_{\mathrm{err}}$
if errors cancel across samples. Together, MAD and $\mathrm{SME}_{\mathrm{err}}$
characterize both magnitude and direction of grading bias.

\section{Results}
\label{sec:results}

\noindent\textbf{Dataset scope.} Results are based on $|T| = 103$ validated
prompt--rubric pairs across two domains (Cultural and Linguistics) and two
dialect communities ({Egyptian Arabic: 70, Iraqi Arabic: 33}). Three target models
are evaluated: \texttt{gemini-3.1-pro-\allowbreak preview} (103 unique prompt--response pairs),
Muse Spark (101), and \texttt{grok-4.3} (98), yielding 302 unique
prompt--response pairs graded by human SMEs. Five judge models produce 1,307
judge evaluations.

{
\subsection{SME Ground Truth: Target and Domain Performance}

Table~\ref{tab:sme_domain} shows the mean SME score per target model broken down
by both dialect community and domain. The combined Cultural and Linguistics
split covers both Iraqi Arabic ($N_{\mathrm{IRQ}}=33$: 15 Cultural, 18 Linguistics)
and Egyptian Arabic ($N_{\mathrm{EGY}}=70$: 38 Cultural, 32 Linguistics).

Three findings stand out. First, all three models score substantially higher on
Egyptian than on Iraqi prompts, across both domains. For \texttt{gemini-3.1-pro-\allowbreak preview},
the gap is $+16.67\%$ (IRQ Cultural) vs $+66.75\%$ (EGY Cultural) and $+32.47\%$
(IRQ Linguistics) vs $+95.21\%$ (EGY Linguistics). For Muse Spark the gap is
even more striking: $-23.15\%$ on IRQ Cultural versus $+64.09\%$ on EGY Cultural.
Second, \texttt{grok-4.3} scores negative on all Iraqi cells, indicating that the
penalty from triggered negative criteria exceeds the credit from passed positive
criteria on those prompts. Third, the Linguistics scores are consistently higher
than Cultural scores within the same dialect for gemini and Muse Spark, while
grok-4.3 shows the opposite pattern on Egyptian prompts ($+8.40\%$ EGY Linguistics
vs $+19.29\%$ EGY Cultural). This dialect-domain interaction was not visible in a
pure domain breakdown and motivates the separate analysis in
Section~\ref{sec:results:error_distribution}.

The consistent EGY $>$ IRQ score gap is large and robust, holding for all three
target models and within both domains. Its interpretation is complicated by a
confound identified through post-hoc analysis of our SME panel: Egyptian SMEs
grade more leniently than Iraqi SMEs, both in the number and weight of criteria
they author and in the thresholds they apply when assigning PASS/FAIL verdicts.
As a result, the score gap may reflect differences in SME grading stringency
rather than, or in addition to, differences in model knowledge of the two
dialect communities. Separating these contributions requires multi-annotator
ground truth with cross-dialect SME calibration, which is beyond the scope of
the current dataset; we flag this as a future work in
Section~\ref{sec:discussion}.

\begin{table}[H]
\centering
\caption{Mean SME score by target model, dialect community, and domain. Scores
represent the average percentage of rubric points earned, as graded by human SMEs.
$n$ = unique prompt--response pairs in that cell. Domain is assigned at the prompt
level via an explicit field in the dataset (Cultural or Linguistics); both domain
values span both dialect communities.}
\label{tab:sme_domain}
\small
\renewcommand{\arraystretch}{1.35}
\begin{tabularx}{\textwidth}{l C C C C C}
  \toprule
  & & \multicolumn{2}{c}{\textbf{Iraqi Arabic (IRQ)}}
    & \multicolumn{2}{c}{\textbf{Egyptian Arabic (EGY)}} \\
  \cmidrule(lr){3-4} \cmidrule(lr){5-6}
  \textbf{Target Model} & \textbf{Pairs}
    & \textbf{Cultural} & \textbf{Linguistics}
    & \textbf{Cultural} & \textbf{Linguistics} \\
  \midrule
  \rowcolor{rowgray}
  gemini-3.1-pro-preview & 103
    & $+16.67\%$ ($n=15$) & $+32.47\%$ ($n=18$)
    & $+66.75\%$ ($n=38$) & $+95.21\%$ ($n=32$) \\
  Muse Spark & 101
    & $-23.15\%$ ($n=15$) & $-22.68\%$ ($n=18$)
    & $+64.09\%$ ($n=36$) & $+72.81\%$ ($n=32$) \\
  \rowcolor{rowgray}
  grok-4.3 & 98
    & $-9.39\%$ ($n=15$)  & $-7.86\%$ ($n=18$)
    & $+19.29\%$ ($n=33$) & $+8.40\%$ ($n=32$) \\
  \midrule
  \textbf{Overall} & 302
    & \multicolumn{4}{c}{Human SME mean: $+36.17\%$} \\
  \bottomrule
\end{tabularx}
\end{table}
}

\subsection{Judge Performance}
\label{sec:results:judge_performance}

\begin{figure}[H]
  \centering
  \includegraphics[width=0.80\textwidth]{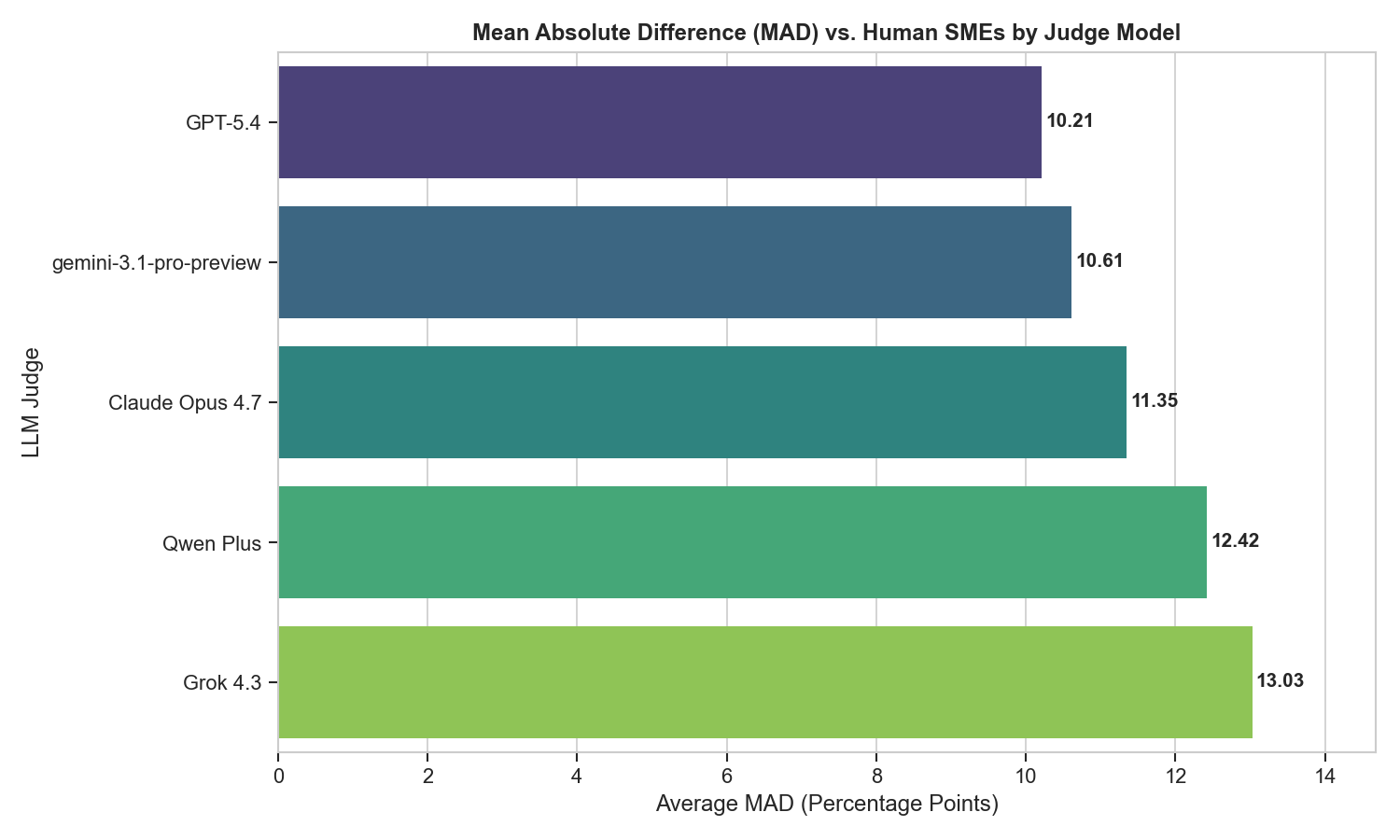}
  \caption{%
    \textbf{$\mathrm{MAD}_j$ by Judge Model.}
    Mean Absolute Difference vs.\ human SME across 302 unique prompt--response pairs.
    GPT-5.4 achieves the lowest MAD (10.21~pp) with near-zero directional bias
    ($-1.12\%$, slightly conservative). Grok~4.3 shows the highest MAD (13.03~pp).
    All judges except GPT-5.4 are lenient (positive $\mathrm{SME}_{\mathrm{err}}$).
  }
  \label{fig:mad_bar}
\end{figure}

\begin{figure}[H]
  \centering
  \includegraphics[width=0.90\textwidth]{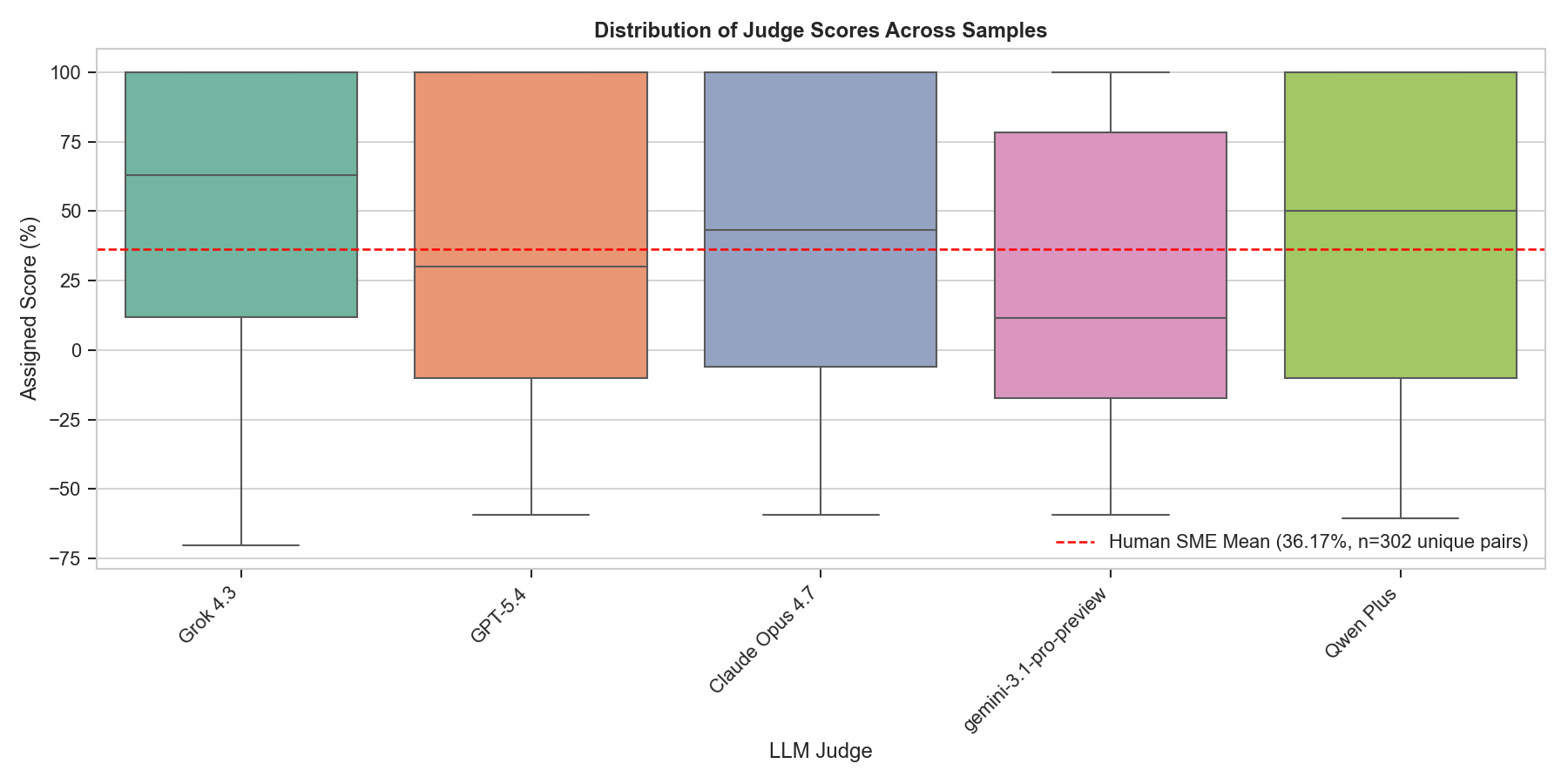}
  \caption{%
    \textbf{Distribution of Judge Scores Across Samples.}
    Box plots show the full distribution of judge-assigned scores per
    prompt--response pair. The red dashed line marks the human SME mean
    ($+36.17\%$ after excluding pilot-only target models).
    Grok~4.3 has the widest IQR and median furthest above the SME line,
    consistent with highest MAD and positive $\mathrm{SME}_{\mathrm{err}}$.
    GPT-5.4 has the most compact distribution.
  }
  \label{fig:boxplot}
\end{figure}

Table~\ref{tab:judge_summary} reports $\mathrm{MAD}_j$ and
$\mathrm{SME}_{\mathrm{err}}$ for each judge across all 302 unique
prompt--response pairs. { Figure~\ref{fig:mad_bar} plots the five
$\mathrm{MAD}_j$ values in ascending order; Figure~\ref{fig:boxplot} shows the
full distribution of judge-assigned scores, with the human SME mean
($+36.17\%$) marked as a reference line.

\textbf{GPT-5.4 is the most reliable judge considering the SME judges as refernce.} It achieves the lowest
$\mathrm{MAD}_j$ (10.21~pp) and the only conservative bias
($\mathrm{SME}_{\mathrm{err}} = -1.12\%$): it occasionally
under-credits responses the SME would pass, but it does not systematically
over-credit responses --- the more dangerous failure mode in a quality-control
setting where the goal is to identify errors. Its score distribution
(Figure~\ref{fig:boxplot}) is the most compact across all five judges, with
the smallest inter-quartile range and a median closest to the SME line.

\textbf{Four of five judges exhibit systematic leniency.}
$\mathrm{SME}_{\mathrm{err}}$ is positive for \texttt{gemini-3.1-pro-\allowbreak preview}
($+3.32\%$), Claude Opus~4.7 ($+6.10\%$), Qwen Plus ($+6.56\%$), and
Grok~4.3 ($+2.01\%$). Leniency is consequential on a rubric dominated by
negative criteria: a lenient judge does not merely assign a slightly higher
score --- it systematically underestimates the error rate of target responses,
which is the primary signal the rubric is designed to capture. Grok~4.3 is an
interesting outlier: it shows the smallest positive $\mathrm{SME}_{\mathrm{err}}$
among the lenient judges ($+2.01\%$) yet the highest $\mathrm{MAD}_j$
(13.03~pp), and its score distribution (Figure~\ref{fig:boxplot}) has the
widest IQR and median furthest above the SME reference line. High MAD with
low directional bias indicates erratic rather than systematically biased
grading: Grok~4.3 swings both above and below the SME on individual samples,
with the net error cancelling to a small positive mean. The direction of error
therefore understates its unreliability as a judge.}

\begin{table}[H]
\centering
\caption{Judge model performance. $\mathrm{SME}_{\mathrm{err}} > 0$: lenient
(scores above SME). $\mathrm{SME}_{\mathrm{err}} < 0$: conservative. Human SME
mean: $+36.17\%$ across 302 unique prompt--response pairs.}
\label{tab:judge_summary}
\small
\renewcommand{\arraystretch}{1.35}
\begin{tabularx}{\textwidth}{l C C C L}
  \toprule
  \textbf{Judge Model}
    & \textbf{$\mathrm{MAD}_j$ (pp)}
    & \textbf{$\mathrm{SME}_{\mathrm{err}}$ (\%)}
    & \textbf{Eval Rows}
    & \textbf{Verdict} \\
  \midrule
  \rowcolor{rowgray}
  \textbf{GPT-5.4}
    & \textbf{10.21} & $-1.12$ & 302 & \textbf{Best judge} \\
  gemini-3.1-pro-preview
    & 10.61 & $+3.32$ & 199 & Reliable \\
  \rowcolor{rowgray}
  Claude Opus 4.7
    & 11.35 & $+6.10$ & 302 & Reliable \\
  Qwen Plus
    & 12.42 & $+6.56$ & 300 & Reliable \\
  \rowcolor{rowgray}
  Grok 4.3
    & 13.03 & $+2.01$ & 204 & Highest MAD \\
  \bottomrule
\end{tabularx}
\vspace{3pt}\\
{\footnotesize Note: gemini and Grok~4.3 have fewer evaluation rows (199, 204)
than the other judges because, as target models themselves, all same-provider
pairs are excluded by the provider-level self-evaluation guard. Qwen Plus
has 300 rather than 302 rows because its content filter declined to evaluate
two Muse Spark responses; those pairs
are excluded from Qwen Plus's MAD computation.}
\end{table}

\subsection{Target-Axis Analysis}

Table~\ref{tab:target_mad} shows the target-axis MAD --- how consistently all
judges agree with the SME when grading a specific target's responses. A high value
indicates responses that are hard to grade automatically, independent of which
judge is used.

{
\textbf{Response quality and gradeability do not scale together.}
Muse Spark, the middle-quality target ($+38.43\%$), is the easiest to grade
(Target MAD 10.05~pp). Both the strongest target ---
\texttt{gemini-3.1-pro-\allowbreak preview} ($+62.31\%$, MAD 12.00~pp) --- and the
weakest --- \texttt{grok-4.3} ($+6.36\%$, MAD 12.77~pp) --- are harder to
grade than Muse Spark. The reason differs at each extreme: strong responses
require judges to credit nuanced correct answers they tend to under-value,
while weak responses tempt judges to over-credit the few surface elements
that are correct. Muse Spark's errors, by contrast, are explicit enough that
all judges catch them reliably. Target MAD therefore measures how
\emph{interpretable} a response is to an automated judge, not how good it
is. The mechanisms behind this pattern are examined further in Section~\ref{sec:discussion:high-quality-responses}.
}

\begin{table}[H]
\centering
\caption{Target model performance. Target MAD = average deviation across all
cross-provider judges when grading that target's responses.}
\label{tab:target_mad}
\small
\renewcommand{\arraystretch}{1.35}
\begin{tabularx}{\textwidth}{l C C C}
  \toprule
  \textbf{Target Model} & \textbf{Unique Pairs} & \textbf{Mean SME Score (\%)} & \textbf{Target MAD (pp)} \\
  \midrule
  \rowcolor{rowgray}
  gemini-3.1-pro-preview & 103 & $+62.31$ & 12.00 \\
  Muse Spark             & 101 & $+38.43$ & 10.05 \\
  \rowcolor{rowgray}
  grok-4.3               &  98 & $+6.36$  & 12.77 \\
  \bottomrule
\end{tabularx}
\vspace{3pt}\\
{\footnotesize \textit{Unique Pairs} = unique (sample, target) pairs graded by human
SMEs; one row per pair regardless of how many judges evaluated it.}
\end{table}

\subsection{Domain Breakdown: Cultural vs.\ Linguistics}
\label{sec:results:domain_breakdown}

Table~\ref{tab:domain} shows $\mathrm{MAD}_j$ split by domain for each judge.
All five judges show higher MAD on Cultural tasks than on Linguistics tasks (negative
gap), indicating that implicit cultural reasoning --- social conventions, contextual
appropriateness, dialect norms --- presents a harder target for automated judges than
linguistically verifiable criteria.

{
\textbf{The domain gap is universal but varies in magnitude.}
Grok~4.3 shows the largest gap ($-4.78$~pp: Cultural MAD 15.37 vs.\
Linguistics MAD 10.59) and GPT-5.4 the second largest ($-4.00$~pp: 12.20
vs.\ 8.20). At the other end, Qwen Plus shows the smallest gap
($-1.83$~pp: 13.33 vs.\ 11.50), and Claude Opus~4.7 similarly modest
($-2.60$~pp). The sign is negative for all five judges; no judge is
better at Cultural than Linguistics grading. Because the domain assignment
is made at the prompt level and the same rubric structure (positive
and negative criteria, explicit and implicit tags) is used throughout,
the gap cannot be attributed to rubric format differences between domains.
It reflects instead the qualitative difference in what Cultural criteria
require: a judge must simulate native-dialect perspective and assess whether
a model's cultural reference is authentic --- a form of open-ended reasoning
with no lexical check available --- whereas Linguistics criteria typically
ask the judge to verify specific morphological forms, syntactic constructions,
or register norms against enumerable, verifiable criteria. This maps onto
the Explicit vs.\ Implicit criterion taxonomy in the rubric
(Section~\ref{sec:dataset:rubric}) and is examined further in
Section~\ref{sec:discussion:cultural_gap}.
}

\begin{table}[H]
\centering
\caption{MAD by domain and judge model. Gap = Linguistics MAD $-$ Cultural MAD.
All negative values confirm Cultural tasks are harder to grade automatically.}
\label{tab:domain}
\small
\renewcommand{\arraystretch}{1.35}
\begin{tabularx}{\textwidth}{l C C C}
  \toprule
  \textbf{Judge Model} & \textbf{Cultural MAD (pp)} & \textbf{Linguistics MAD (pp)} & \textbf{Gap (pp)} \\
  \midrule
  \rowcolor{rowgray}
  GPT-5.4                & 12.20 & 8.20  & $-4.00$ \\
  gemini-3.1-pro-preview & 12.04 & 9.20  & $-2.84$ \\
  \rowcolor{rowgray}
  Claude Opus 4.7        & 12.64 & 10.04 & $-2.60$ \\
  Qwen Plus              & 13.33 & 11.50 & $-1.83$ \\
  \rowcolor{rowgray}
  Grok 4.3               & 15.37 & 10.59 & $-4.78$ \\
  \bottomrule
\end{tabularx}
\end{table}

{
\subsection{Error-Type Distribution Across the Full Dataset}
\label{sec:results:error_distribution}

This section characterises \emph{what kinds of errors} the three target models
actually make and how their profiles differ between dialect communities. Every
failed criterion is tagged with one of the six failure types from
Section~\ref{sec:dataset:rubric}, counted across all three included targets
(\texttt{gemini-3.1-pro-\allowbreak preview}, Muse Spark, \texttt{grok-4.3}).

\textbf{Analysis design.} Because Egyptian SMEs grade more leniently than Iraqi
SMEs --- particularly for \textsc{Cultural Misattribution} and \textsc{Ambiguous
Framing} (see Section~\ref{sec:discussion:errortypes}) --- the full
six-error-type breakdown is presented for Iraqi Arabic only. Egyptian Arabic data
are included exclusively for \textsc{Hallucination}, the most objective error type
(a fabricated fact is verifiably wrong regardless of SME stringency), which serves
as a cross-dialect calibration anchor.

\textbf{Rate metric.} Because three models each evaluate every prompt, normalising
by raw prompt count inflates rates by a factor of three. All rates therefore use
\textbf{model-prompt evaluation pairs} as the denominator:
$\text{Rate} = \dfrac{\text{error count}}{\text{model-prompt pairs}} \times 100$.
Iraqi Arabic has $99$ pairs ($3 \times 33$; perfect coverage); Egyptian Arabic has
$203$ pairs (Muse Spark: 68; \texttt{gemini-3.1-pro-\allowbreak preview}: 70;
\texttt{grok-4.3}: 65), giving ${\approx}2.9$ evaluations per prompt --- nearly
balanced with Iraqi ($3.0$). For example, Muse Spark's 74 Cultural
Misattribution errors across 33 Iraqi pairs give a per-model rate of $224.2$;
pooling all three models: $172/99 \times 100 = 173.7$ per 100 pairs.

\textbf{Iraqi error profile and per-model breakdown.}
\textsc{Cultural Misattribution} and \textsc{Ambiguous Framing} lead at 172
instances each, followed by \textsc{Omission} (145), \textsc{Hallucination} (140),
\textsc{Irrelevant Addition} (38), and \textsc{Rendering Failure} (0). The
per-model breakdown (Figure~\ref{fig:err_fig5v2}) reveals a strong dependence on
target quality. \texttt{grok-4.3}, the weakest target, is dominated by
\textsc{Omission} (76): it most often fails to produce the correct term entirely.
Muse Spark records the highest \textsc{Cultural Misattribution} (74) and
\textsc{Ambiguous Framing} (73) while keeping \textsc{Omission} lower (47): it
surfaces the term but wraps it in misattributed or register-inconsistent content.
\texttt{gemini-3.1-pro-\allowbreak preview}, the strongest target, has the lowest
\textsc{Omission} (22), with residual errors in \textsc{Ambiguous Framing} (55)
and \textsc{Irrelevant Addition} (20). Strikingly, \texttt{grok-4.3} inverts its
profile across dialects: \textsc{Omission}-dominated on Iraqi prompts (76) but
\textsc{Hallucination}-dominated on Egyptian (106), fabricating confidently where
it fell silent on Iraqi. The cross-dialect interpretation of this contrast is
deferred to Section~\ref{sec:discussion:errortypes}.

\begin{figure}[H]
  \centering
  \includegraphics[width=0.95\textwidth]{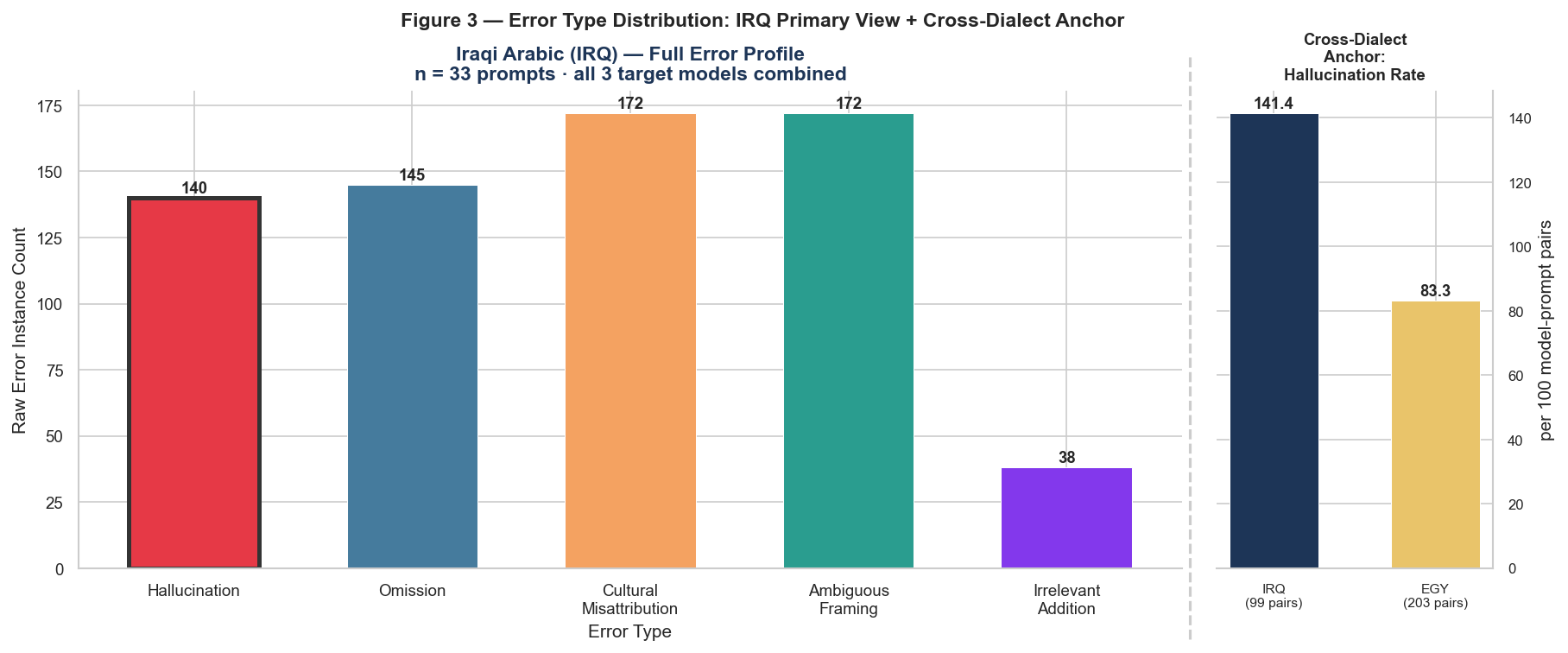}
  \caption{%
    \textbf{Aggregated Iraqi Error Profile and Hallucination Cross-Dialect Anchor.}
    \emph{Left panel:} Raw error counts for Iraqi Arabic across all three included
    targets.
    \emph{Right panel:} \textsc{Hallucination} rate per 100 model-prompt pairs.
  }
  \label{fig:err_fig3v2}
\end{figure}

\begin{figure}[H]
  \centering
  \includegraphics[width=0.95\textwidth]{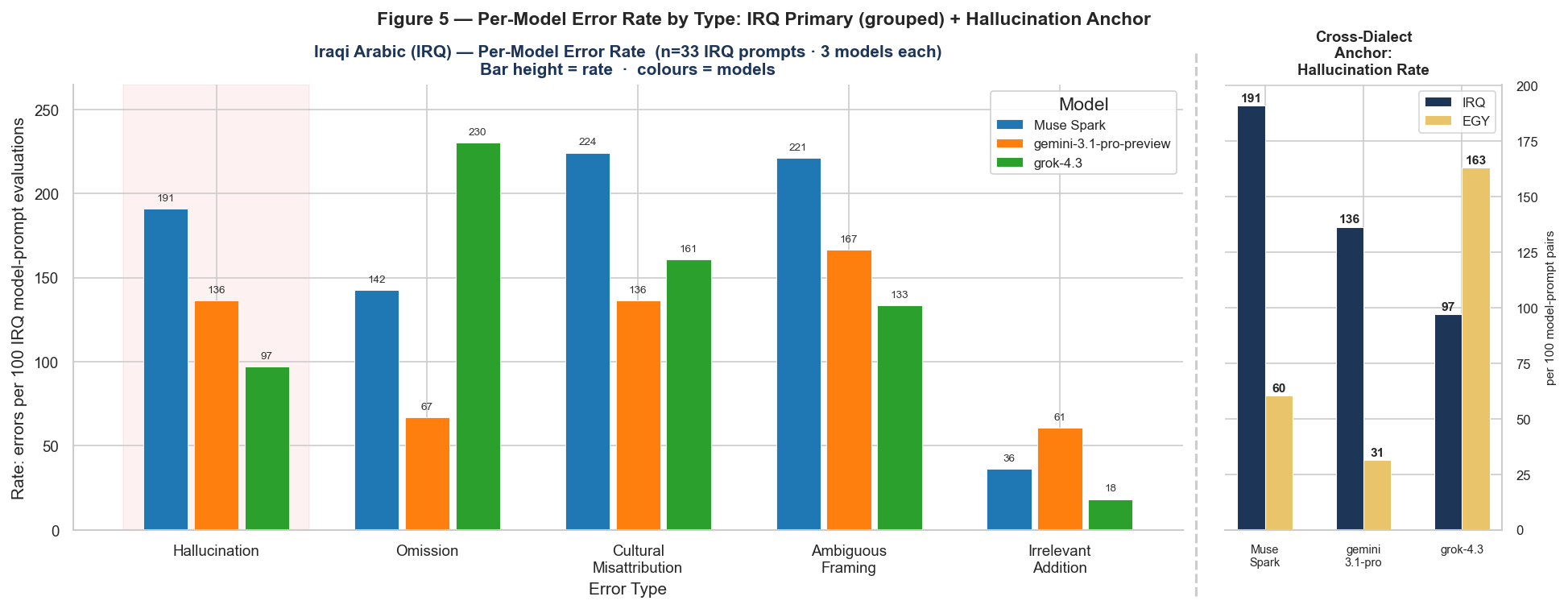}
  \caption{%
    \textbf{Per-Model Iraqi Error Profile and Hallucination Anchor.}
    \emph{Main panel:} Iraqi error counts for all six types, with the three target
    models clustered per group.
    \emph{Anchor panel (right):} \textsc{Hallucination} rate per 100 model-prompt
    pairs by model for Iraqi (navy) and Egyptian (gold).
  }
  \label{fig:err_fig5v2}
\end{figure}

\begin{figure}[H]
  \centering
  \includegraphics[width=0.95\textwidth]{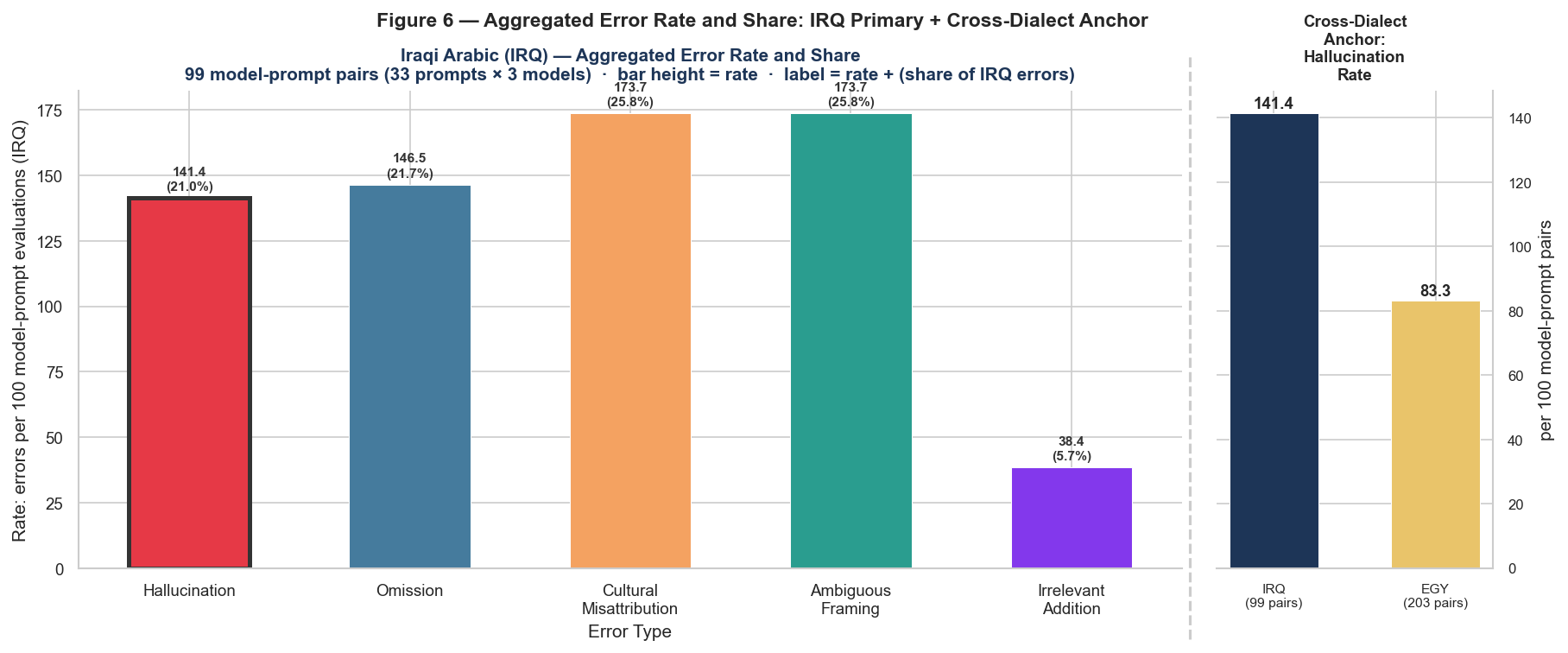}
  \caption{
    \textbf{Aggregated Iraqi Error Rate and Share, with Hallucination Anchor.}
    \emph{Left panel:} Rate (errors per 100 model-prompt evaluations; IRQ denominator
    $= 99$ pairs) and share for Iraqi Arabic. \textsc{Cultural Misattribution} and
    \textsc{Ambiguous Framing} are highest (173.7 each), followed by \textsc{Omission}
    (146.5) and \textsc{Hallucination} (141.4).
    \emph{Right panel:} \textsc{Hallucination} anchor --- IRQ vs.\ EGY
    per 100 model-prompt pairs.
  }
  \label{fig:err_fig6v2}
\end{figure}


\section{Discussion}
\label{sec:discussion}

\textbf{Judge reliability and the leniency failure mode.}\label{sec:discussion:cultural_gap}
GPT-5.4 is the most reliable judge (MAD 10.21~pp, Signed Error $-1.12\%$); its
slight conservative bias is the safer failure mode in a quality-control setting.
Four of five judges are lenient ($+2.01\%$--$+6.56\%$), which is consequential
on a penalty-dominated rubric: leniency underestimates error rates, the primary
signal the rubric captures. The consistent Cultural $>$ Linguistics MAD gap
($1.83$--$4.78$~pp, Table~\ref{tab:domain}) reflects an Implicit criterion
requirement: Cultural tasks demand native-dialect simulation rather than lexical
verification, a distinction that maps onto the Explicit vs.\ Implicit taxonomy
of Section~\ref{sec:dataset:rubric}.

\textbf{Why Egyptian error rates look so much lower --- and why we cannot take that at face value.}\label{sec:discussion:errortypes}
Two error types stand out in the Iraqi data: \textsc{Cultural Misattribution} and
\textsc{Ambiguous Framing} occur roughly $117\times$ and $39\times$ more often per
evaluation in the Iraqi subset than in the Egyptian subset. The intuitive explanation
is that models know Egyptian Arabic better --- it is more prevalent in pretraining
data, so models are less likely to substitute a wrong dialect term. That may well be
part of the story.

However, our SME panel reveals a second explanation that we cannot rule out:
Egyptian SMEs write fewer negative criteria per response and apply more lenient
PASS/FAIL thresholds than Iraqi SMEs. \textsc{Cultural Misattribution} and
\textsc{Ambiguous Framing} are precisely the error types that a lenient grader is
most likely to leave undocumented --- catching them requires noticing a subtle
dialect mismatch and then deciding to record it as a penalty. A strict grader
documents it; a lenient one may let it pass.

To test which explanation is more likely, we use \textsc{Hallucination} as a
reference. A hallucinated fact is objectively wrong regardless of how strict the
grader is --- there is no leniency involved in recognising that a model invented a
term or attributed a custom to the wrong region. If the gap between Iraqi and
Egyptian were driven mainly by genuine model-knowledge differences, we would expect
\textsc{Hallucination} to show a similarly large gap. Instead, \textsc{Hallucination}
shows only a $1.7\times$ IRQ/EGY ratio (Figure~\ref{fig:err_fig6v2}), the smallest
of any error type. The fact that the one objective, leniency-resistant error type
shows a near-equal split while the two judgment-dependent types show $100\times$+
gaps is consistent with SME grading stringency --- not model knowledge --- driving
much of the observed difference. We therefore withhold causal attribution to model
behaviour; separating the two contributions requires a cross-dialect SME calibration
study.

\textbf{Being right and verbose can score lower than being right and brief.}\label{sec:discussion:high-quality-responses}
Looking at Target MAD, the middle-quality model is actually the easiest to grade:
Muse Spark ($+38.43\%$ mean SME score, MAD 10.05~pp) is graded more consistently
than either \texttt{gemini-3.1-pro-\allowbreak preview} ($+62.31\%$, MAD 12.00~pp)
or \texttt{grok-4.3} ($+6.36\%$, MAD 12.77~pp). The reason is not that Muse Spark
is average --- it is in fact the best of the three at identifying the correct dialect
term, often finding the gold answer on prompts where the other models miss it
entirely. The problem is what it does next. Rather than stopping, Muse Spark
routinely elaborates: it adds alternative terms, invented etymologies, regional
usage claims, and unsolicited background. Each extra clause is a new opportunity to
trigger a negative criterion --- a fabricated detail fires \textsc{Hallucination}, a
non-native phrasing fires \textsc{Ambiguous Framing}, an off-topic addition fires
\textsc{Irrelevant Addition} or \textsc{Cultural Misattribution}. The penalty mass
from this elaboration erodes the credit earned by the correct identification. In a
penalty-weighted rubric, being right and long can score below being right and short:
correctness is bounded by fixed positive weights, while error opportunity grows with
response length.

This also explains why Muse Spark posts the \emph{lowest} Target MAD despite its
strong recall. Its errors --- Cultural Misattribution (74), Ambiguous Framing (73)
--- are explicit and lexical; every judge can see them and agrees on them. By
contrast, \texttt{grok-4.3}'s dominant failure is \textsc{Omission} (76 on Iraqi):
it simply does not produce the correct term, giving judges little to disagree about
either. \texttt{gemini}'s high-quality but nuanced responses create the opposite
problem --- judges under-credit elaborations that the SME would accept. The three
models therefore represent three distinct error geometries, not three points on a
single quality axis, which is why Target MAD does not scale monotonically with
response quality.

\subsection{Limitations}

\begin{itemize}
  \item \textbf{Single SME per sample.} Each (target model, sample) pair is graded
        by a single SME, leaving annotator noise unmeasured. Inter-annotator
        agreement is the empirical ceiling for any automated judge and cannot be
        quantified from the current dataset.
  \item \textbf{Differential SME grading stringency.} Post-hoc analysis of our SME
        panel indicates that Egyptian SMEs grade more leniently than Iraqi SMEs,
        both in the number and severity of criteria they author and in the PASS/FAIL
        thresholds they apply. This confounds all cross-dialect comparisons: the
        observed EGY $>$ IRQ performance gap and the concentration of
        \textsc{Cultural Misattribution} and \textsc{Ambiguous Framing} in the
        Iraqi subset may reflect SME grading behaviour as much as model behaviour.
        We cannot separate these contributions from the current data.
  \item \textbf{Two-dialect coverage.} The dataset covers Egyptian and Iraqi Arabic
        only. Findings may not generalise to other MENA dialect communities.
\end{itemize}

\subsection{Future Work}

\begin{itemize}
  \item \textbf{Multi-annotator grading.} Recruiting two or more SMEs per sample
        would establish inter-annotator agreement as an empirical ceiling and allow
        the annotator-noise component of judge MAD to be isolated.
  \item \textbf{Cross-dialect SME calibration.} A calibration study in which SMEs
        from both dialect communities grade the same subset of responses would
        quantify grading-stringency differences and allow the SME-leniency confound
        to be separated from genuine model-knowledge differences.
  \item \textbf{Dialect expansion.} Extending the benchmark to Jordanian, Gulf, and
        Levantine Arabic requires additional SME recruitment and data collection but
        would enable cross-dialect generalisability claims.
\end{itemize}

\section{Conclusion}
\label{sec:conclusion}

We have presented a cross-evaluation framework for benchmarking frontier LLMs as
automated graders in Arabic cultural and sociolinguistic domains, evaluated on a
dataset of 103 validated prompt--rubric pairs across Cultural and Linguistics domains
in Egyptian and Iraqi Arabic. The framework's provider-level self-evaluation guard,
penalty-weighted rubric design, and dual-metric reporting (MAD + Signed Mean Error)
are designed to surface specific failure modes of automated grading rather than
produce a single aggregate score.

The key findings are: GPT-5.4 is the most reliable automated judge (MAD 10.21~pp,
near-zero bias); four of five judges exhibit systematic leniency
($+2.01\%$--$+6.56\%$); Cultural tasks are harder to grade automatically than
Linguistics tasks for all five judges (gap $1.83$--$4.78$~pp); and the difficulty of
grading a target's responses does not scale monotonically with response quality ---
both high-quality (gemini) and low-quality (grok-4.3) responses are harder to
grade than moderate-quality ones (Muse Spark).

Expanding the dataset to additional MENA dialect communities,
{calibrating SME grading stringency across dialect communities
through multi-annotator cross-dialect protocols,} introducing multi-annotator
SME grading to establish inter-annotator agreement as an empirical ceiling, and
investigating the Explicit vs.\ Implicit criterion gap at the per-criterion
level are the natural next steps. The answer to the question of whether
any frontier LLM can reliably substitute for a domain-expert SME across the full
range of Arabic cultural knowledge tasks has direct practical implications for the
scalability of human-expert evaluation in Arabic NLP and for the deployment of LLMs
in culturally sensitive contexts.

\bibliographystyle{plain}
\bibliography{references}

\section{Judge Prompt}
\label{app:judge_prompt}

All judge calls use the following fixed system prompt and user message template.
Temperature is set to 0 for all calls; no other sampling parameters are varied.

\subsection*{System Prompt}

\begin{quote}\ttfamily\small
You are an expert Subject Matter Expert (SME) in Arabic Sociolinguistics and
Dialects. You will be provided with a PROMPT, an ANSWER generated by an AI model,
and a grading RUBRIC. You must evaluate the ANSWER against every criterion in the
RUBRIC.

Return ONLY a JSON object with the following structure:

\{

\quad ``evaluations'': [

\qquad \{

\qquad\quad ``criterion'': ``Text of criterion here'',

\qquad\quad ``score\_type'': ``Positive or Negative'',

\qquad\quad ``judgment'': ``PASS or FAIL (if Positive), or 0 or Error Present (if Negative)'',

\qquad\quad ``justification'': ``Your reasoning here''

\qquad \}

\quad ]

\}
\end{quote}

\subsection*{User Message Template}

\begin{quote}\ttfamily\small
PROMPT: \textit{\{prompt text authored by SME\}}

---

ANSWER (produced by \textit{\{target model name\}}): \textit{\{target model's response\}}

---

RUBRIC:

\textit{\{plain-text table: Criterion $\mid$ Score Type $\mid$ Weight, one row per criterion\}}
\end{quote}

\noindent The rubric table contains three columns only: criterion text, score type
(\texttt{Positive} or \texttt{Negative}), and numeric weight. The SME's
PASS/FAIL verdicts and failure-type tags are deliberately excluded so the judge
receives no ground-truth signal. The target model's name is included in the
user message to allow attribution in the judge's justification text; it is
not used for scoring. Judgment tokens (\texttt{PASS}, \texttt{FAIL},
\texttt{Error Present}, \texttt{0}) are parsed by exact string match in the
evaluation pipeline.
}

\end{document}